\documentclass[runningheads]{llncs}
\usepackage[T1]{fontenc}
\usepackage{graphicx}
\usepackage{minted}
\usepackage{amsmath}
\usepackage{hyperref}
\usepackage{color}

\begin{document}
\title{PCRLLM: Proof-Carrying Reasoning with Large Language Models under Stepwise Logical Constraints}
\titlerunning{PCRLLM}

%
%
\author{Tangrui Li\orcidID{0000-0003-3110-8324} \and
Pei Wang \and
Hongzheng Wang \and
Christian Hahm \and
Matteo Spatola \and
Justin Shi\orcidID{0000-0003-3774-2190} }
%
\authorrunning{T. Li, P. Wang, et al.}
%
\institute{Temple University, Philadelphia PA 19122, USA \\
\email{\{tuo90515,pei.wang,tuf78197,tun50806,matteo.spatola, shi\}@temple.edu}}
\maketitle              
\begin{abstract}

Large Language Models (LLMs) often exhibit limited logical coherence, mapping premises to conclusions without adherence to explicit inference rules. We propose Proof-Carrying Reasoning with LLMs (PCRLLM), a framework that constrains reasoning to single-step inferences while preserving natural language formulations. Each output explicitly specifies premises, rules, and conclusions, thereby enabling verification against a target logic. This mechanism mitigates trustworthiness concerns by supporting chain-level validation even in black-box settings. Moreover, PCRLLM facilitates systematic multi-LLM collaboration, allowing intermediate steps to be compared and integrated under formal rules. Finally, we introduce a benchmark schema for generating large-scale step-level reasoning data, combining natural language expressiveness with formal rigor.


\end{abstract}
\section{Introduction}

The reasoning ability of large language models (LLMs) has long been a central concern, yet their answers often lack stable logical coherence. This stems from their nature as neural networks, which treat reasoning as a direct mapping from premises to conclusions according to the training data. Such mapping is not inherently problematic, but it should be constrained by logical rules and restricted to the smallest unit of inference (as single-step reasoning). Otherwise, when LLMs skip steps by leveraging semantic knowledge, we cannot tell whether they are performing valid logical short-cuts or unjustified leaps (thus losing the soundness). To this point, having LLMs simulate an existing logical system is a reasonable choice, since the rules of single-step inference are explicitly defined.

Nevertheless, LLMs' strength lies not in mechanically reproducing logic systems, but in their powerful natural language processing ability. Existing approaches usually emphasize only one side: either broad reasoning ability without formal soundness (e.g., CoT \cite{CoT}), or strict formalism at the cost of generality (e.g., neural-symbolic architectures \cite{neuralsymbolic}, domain-specific loss on formality \cite{chemistry}). Our approach aims to combine the two: retaining natural language problem descriptions while rephrasing them in formal logical reasoning in the output, which is a chain reasoning consists of well-formed single steps.

This design not only addresses the above dilemma but also brings two additional advantages. First, it produces proof-carrying outputs: the LLM’s answers include premises, rules, conclusions, and necessary details, which allows each single-step inference to be checked against the target logic, enabling a chain-level grading as an automatic judgment of logical soundness. In this way, the trustworthiness problem can be mitigated: even if the model remains a black box, its outputs can still be validated on logical grounds.

Secondly, it enables more systematic multi-LLM collaboration. Prior approaches mostly rely on debates or answer aggregation, which lack formal constraints and obscure the quality of intermediate steps. In our framework, multiple models’ reasoning steps can be unified through logical rules, so that collaboration improves not only the final result but also the transparency and consistency of intermediate reasoning.

Additionally, we also recognize that the complexity of long-chain reasoning arises from the exponential growth of rule applications over shared premises, which cannot be captured by existing corpora alone. To address this, we propose a benchmark schema that generates large-scale training data, allowing LLMs to better learn complex inference patterns, as well as the testing data.

To summarize, we propose a methodology that enables LLMs to perform formal reasoning while preserving natural language problem descriptions. The key idea is to constrain reasoning to single steps so that each step can be verified against a target logic. This yields proof-carrying outputs that allow automatic scoring of reasoning chains and provide a new angle to address the trustworthiness problem. Moreover, by expressing reasoning under formal constraints, our approach supports systematic multi-LLM collaboration, where intermediate steps can be compared and integrated through logical rules rather than opaque debates. To further overcome the data scarcity of formal reasoning, we introduce a benchmark schema that generates large-scale step-level training data. Together, these contributions offer a principled path toward combining LLMs’ natural language understanding with the rigor of formal logic. All the codes and appendix are in our \href{https://github.com/MoonWalker1997/rLLMFT}{GitHub repository}.

The remainder of this paper 1) reviews related work, analyzes the scope of our approach, and 2) introduces the logical system we adopt along with its necessary background. We then 3) describe our data generation strategy, evaluation of reasoning, and multi-LLM integration, followed by 4) experiments that demonstrate the effectiveness of our method and discuss the challenges of simulating reasoning with LLMs.

\section{Related Work}

\subsection{Approach Introducing Logic to LLMs}

\begin{itemize}

    \item \textbf{CoT and PRM} \\ CoT (Chain of Thought \cite{CoT}) was the first approach to explicitly introduce step-by-step reasoning in LLMs. Its core idea is “reasoning granularity”: prompting models to generate intermediate steps and leveraging techniques such as self-consistency sampling or process rewards to boost performance. However, the underlying mechanism remains closer to semantic pattern mapping than to fully formal reasoning. Even with explicit steps, the transitions lack the strict constraints of a logical system. As a result, while CoT improves performance on arithmetic and logical tasks \cite{CoTmathlogic}, its intermediate reasoning often suffers from faithfulness issues \cite{CoTfaithfulness}. PRM (Process Reward Model \cite{PRM}) shares some similarity with our approach: it evaluates reasoning steps rather than final answers. Yet the reasoning process remains a black box—we can only score or select from the outputs, not reshape them. Moreover, PRM makes progress by evaluating intermediate steps, but its scoring framework is not yet aligned with a formal logical system. As a result, it faces challenges in achieving rigor, generality, or in supporting fully validated chains of evidence.

    \item \textbf{Reasoning with External Tools} \\ The key idea is that LLMs can be nudged toward normative outputs such as code or JSON, exemplified by approaches like PoT \cite{PoT}, where reasoning becomes explicit, step-wise, and programmatic. But these methods leverage programmatic reasoning, which increases explicitness and verifiability, though the reliance on program structures may limit flexibility and complicate cross-model collaboration. A different line of work translates natural language into formal representations and delegates reasoning to external solvers \cite{externalformalsolver}\cite{texttosql}. This, however, faces the semantic ambiguity and uncertainty of natural language \cite{externalformalsolverissues}, and large-scale contexts remain difficult to fully formalize.

    \item \textbf{Neural Symbolic Architecture} \\ This approach encourages neural networks to generate outputs that resemble formal reasoning structures, which can improve logical consistency. In many implementations, however, the reasoning chains are implicit and difficult to validate (with some exceptions \cite{neuralsymbolicoutlier}, though readability and integration into feedback or collaboration remain limited). Furthermore, the underlying logical formalisms are currently focused on propositional or first-order logic \cite{neuralsymbolic}. Since symbolization is closely tied to natural language understanding (NLU), extending these methods to handle unrestricted natural language remains a challenging task.
    
\end{itemize}

\subsection{LLM Collaboration}

LLM collaboration aims to boost performance by combining multiple models. Without a formal structure, however, multi-agent debates \cite{debate1}\cite{debate2} can become subjective. Since natural language is the medium, evaluation often relies on semantic plausibility rather than strict structural validity, which makes guarantees of soundness challenging. Current methods typically fall into two types: 1) expanding reasoning branches via sampling \cite{debate1}, which can be costly and may yield intermediate steps that are difficult to verify; or 2) organizing debates among multiple models \cite{deliberation} (or within the self-talk of a single model \cite{ahamoment}), which may improve accuracy but do not yet provide fully verifiable chains of evidence.

While multi-agent debate has been shown to outperform single models on many tasks \cite{debatehelpful}, it remains constrained by natural language as the carrier: larger debates may yield higher performance \cite{tinyzero}, but also make it harder to automatically parse or validate intermediate reasoning. Benchmarks such as GSM8K \cite{GSM8K} further reinforce this tendency by evaluating only final answers rather than reasoning processes, thus prioritizing correctness of outcomes over explicit scrutiny of intermediate reasoning.

\section{Non-Axiomatic Logic Background}

We choose Non-Axiomatic Logic (NAL) as our target logic for two main reasons. 1) NAL offers sufficient expressive ability to represent natural language contents properly. NAL is defined hierarchically with multiple “layers” and each of them extends the sets of grammar and inference rules of the lower layers, to increase the expressive ability and inferential power of the system\cite{NAL}. The high layers of NAL can support substantially higher semantic complexity. In the meantime, NAL is designed based on AIKR (Assumption of Insufficient Knowledge and Resources) \cite{NAL}, which makes NAL naturally tolerant of the uncertainty and vagueness in natural language. 2) NAL can be considered as a non-monotonic logic. New evidence can strengthen, weaken, or overturn previous judgments through revision. This expands the range of inferences we can draw from LLM outputs, such as allowing different reasoning paths to yield different judgments about the same statement. And more importantly, because of the inherent uncertainty of natural language, judgments from different LLM can rarely share the same truth-value. NAL can naturally accommodate such discrepancies with its non-monotonicity when combining the outputs from multiple LLMs. 

\subsection{Used NAL Rules}

NAL is a complex logic, developed on the basis of syllogistic reasoning and subsequently extended with increasingly powerful expressive capabilities. These capabilities are organized into eight levels of logic rules, with later levels building upon the foundations of earlier ones. For example, NAL-4 introduces acquired relations (including predicates); NAL-5 extends the system to statement-level inference; NAL-6 enables the representation of variables; NAL-7 incorporates temporal descriptions and NAL-8 further expands the system to perception and interaction with the external environment.

In this paper, we restrict our focus to syllogistic inference rules within first-order NAL. To clarify the rules we employ, it is necessary to introduce the following basic concepts:

\begin{itemize}

    \item \textbf{Inheritance} A copula in NAL, denoted as $\rightarrow$. For a judgment $s \rightarrow o$, it indicates that $s$ is a specialization of $o$ (equivalently, $o$ is a generalization of $s$). A low frequency corresponds to the negation of a judgment.
    
    \item \textbf{Similarity} Another copula, denoted as $\leftrightarrow$. $s \leftrightarrow o$ indicates that the two terms can be interchanged as equivalent in the current context. For example, if ``Jack'' first appears in a sentence and is later referred to as ``the one,'' then at that moment the judgment $Jack \leftrightarrow the\ one$ holds true.
    
    \item \textbf{Syllogistic reasoning} Since NAL is founded on sentences of the form ``subject-term copula predicate-term'' and syllogistic reasoning, the application of any rule requires two premises $J_1,J_2$, which must share a common term $M$. The other two terms are denoted as $S$ and $P$.
    
    \item \textbf{Truth-value} A NAL judgment is associated with a two-dimensional truth-value $(f, c)$, where $f$ represents the frequency with which a statement is confirmed based on accumulated evidence, and $c$ represents the confidence, i.e., the stability of $f$. Formally, given equivalent positive evidence $w_p$ and equivalent negative evidence $w_n$, we define $f = w_p/(w_p + w_n)$ and $c = (w_p + w_n)/(w_p + w_n + 1)$. A larger evidence base makes $c$ closer to 1, indicating that $f$ is more stable and therefore more reliable for future reasoning.

    \item \textbf{Evidential base} In NAL, truth-values are determined by evidence, but the system does not track the specific contribution of individual pieces of evidence, as evidence accumulates and changes during inference, making it difficult to trace independently. To prevent double counting, NAL assigns each input statement a unique ID. Any conclusion derived through inference must carry all premise IDs (to a maximum capacity), and if overlapping IDs are detected, the inference is disallowed. The set of accumulated IDs associated with a statement is referred to as its evidential base.
    
    \item \textbf{Truth functions} Since NAL employs continuous truth-values, it is equipped with specially designed functions for calculating them. For instance, $F_{ded}$ is the truth-value function for deduction rule. Since syllogistic inference distinguishes between major and minor premises, the order of the premises matters: the truth-values of $J_1$ and $J_2$ are treated in sequence. When needed, we write $F'_{ded}$ to indicate the function applied after swapping the order of $J_1$ and $J_2$. Besides deduction, we also make use of other rules such as abduction, induction, comparison, analogy, exemplification, and resemblance. For the precise definitions of these rules and their truth functions, please refer to Appendix B in \cite{NAL}.
    
\end{itemize}

The following chart summarizes all the inference rules employed in this paper. This work does not involve an in-depth discussion of NAL. It is sufficient to understand only the basic inference forms required to apply these rules. The table shows which rules can be applied to obtain different conclusions based on different $J_1$ and $J_2$. It is worth noting that NAL often uses all available rules and thus obtains multiple conclusions for a pair of premises.

\begin{figure}
    \centering
    \includegraphics[width=0.75\linewidth]{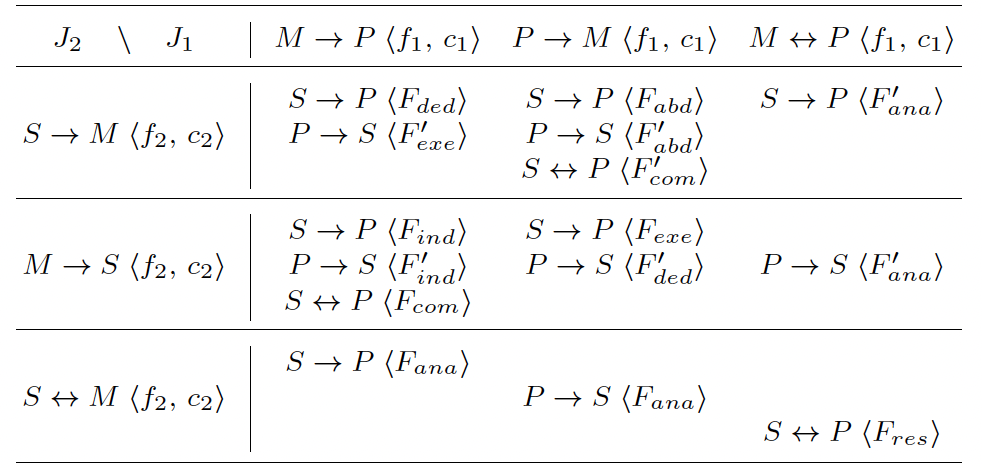}
    \caption{First-order NAL Syllogism.}
    \label{fig:rule table}
\end{figure}

\section{Method}

\subsection{Data Construction}

The data is required to have natural language input as the problem description, and the JSON output as formal NAL inference. In data generation, we will first create a logical structure and then convert it into a sufficiently rich natural language description.

\subsubsection{Logical Instance Generation}

We focus on two-step NAL reasoning in this paper, as it represents the simplest form of multi-step reasoning. Due to computational constraints, we restrict our experiments to this basic case; however, we also emphasize that the proposed method has the potential to be extended to more complex multi-step reasoning tasks when supported by more powerful hardware and larger models.

\begin{itemize}

    \item \textbf{Step 1}: One of the nine inference rules (Fig. \ref{fig:rule table}) is randomly selected to determine first two premises ($J1$, $J2$) used, as well as the pending conclusions. Unique terms (e.g., $ID_{3939}$) are assigned to the placeholders $S$, $M$, $P$ respectively. A random evidential base (within 3 IDs) is attached.
    
    \item \textbf{Step 2}: One conclusion from Step 1 is randomly selected as a premise (J1 or J2), thereby fixing the position of Step 2 in the inference chart. Another independent premise is generated in the same way. The remaining placeholder is assigned a unique term, and one of the possible conclusions of Step 2 is chosen as the final goal.
    
    \item \textbf{Distractors}: To increase difficulty and emphasize the NLU ability of LLMs in such problems, two additional distractor premises are introduced in Step 1, which are generated by randomly selecting a rule and assigning distinct unique terms with random evidential bases.  
    
\end{itemize}

This procedure guarantees that every generated reasoning instance is valid and sufficiently complex. It can be extended to multi-step reasoning or to consider higher levels of NAL rules to further increase complexity.

\subsubsection{Natural Language Instance}

The natural language translation is disclosed to LLMs thus cannot contain everything in the logic instance. It only contains:

\begin{enumerate}
    \item The two premises of Step 1 (without specifying $J1$, $J2$ roles).  
    \item The independent premise of Step 2 (which is not derived from Step 1).  
    \item The two distractor premises for Step 1.  
    \item The selected conclusion of Step 2 (as the question asked to LLMs, truth values removed).  
\end{enumerate}

In translating NAL sentences, 20 templates are prepared for the inheritance copula, as well as the similarity copula. For example, inheritance may be represented as sub is a type of obj; similarity may be represented as sub and obj are conceptually identical. In describing numerical truth-values, the frequency ($f$) is discretized into five categories (\textit{always false, usually false, unknown, usually true, always true}), each with five natural language expressions. Note that the confidence ($c$) is omitted here and replaced with the LLM’s answer-level confidence.  

As a result, each logic instance admits up to $(20 \times 5 \times 5)^5 \times 20$ possible natural language translations, ensuring a high diversity. For all templates used, please refer to the appendix provided in our \href{https://github.com/MoonWalker1997/rLLMFT}{GitHub repository}.

\subsubsection{JSON Representation}

As the expected reponse from LLMs, the step-by-step inference process of each logic instance is represented in JSON format with two top-level keys, which are \textit{Step 1} and \textit{Step 2}. 

In each step, there are 3 keys: \textit{Premise 1}, \textit{Premise 2} and \textit{Results}. In each a NAL sentence $s\ cp\ o. \langle f;c\rangle [eb] \{r\} $ is enclosed with keys: \textit{s} which is the subject; \textit{o} which is the object; \textit{cp} which is the copula; \textit{f} which is the frequency in the truth-value; \textit{c} which is the confidence in the truth-value, here the default value of 0.9 is used; \textit{eb} which is the evidential base of the truth-value; \textit{r} which is the only applicable for sentences in the results, representing the reasoning rule used.

\begin{figure}
     \centering
     \includegraphics[width=0.75\linewidth]{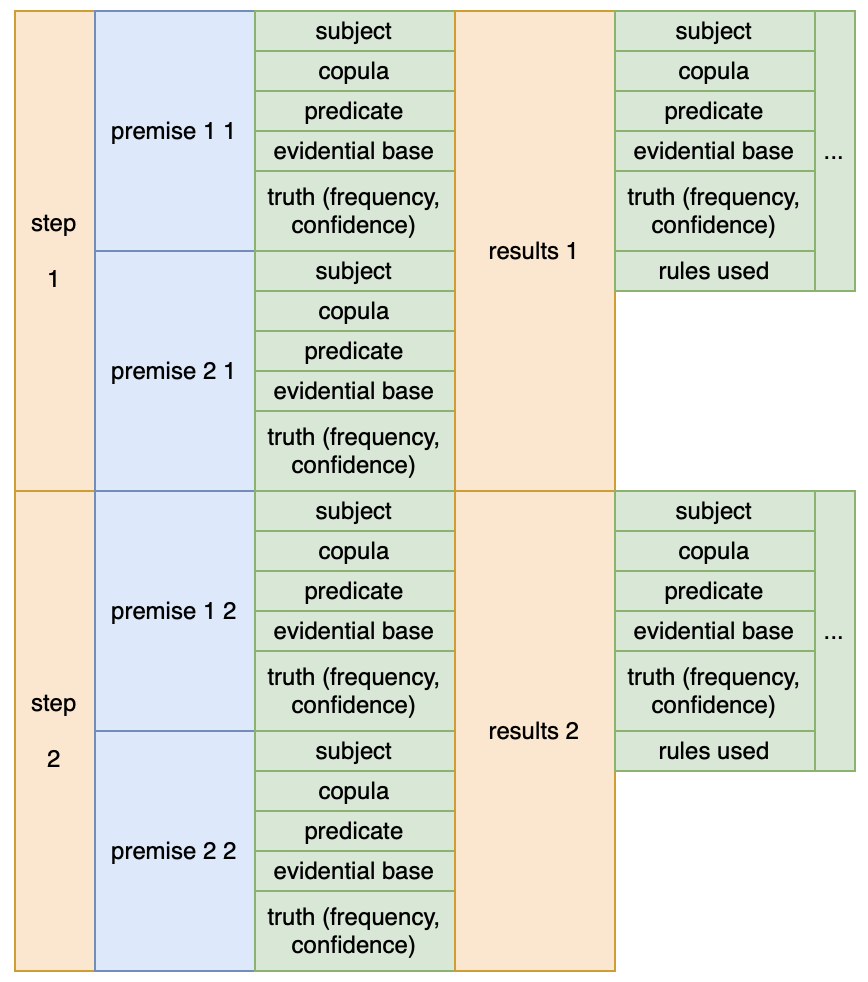}
     \caption{The JSON structure overview of the expected LLM response.}
     \label{fig:JSON structure}
\end{figure}

This structure  preserves the full reasoning process and aligns precisely with the natural language input. The overview of the JSON structure can be found in \ref{fig:JSON structure}.

\subsubsection{Prompt Construction}

\begin{figure}
    \centering
    \includegraphics[width=1.0\linewidth]{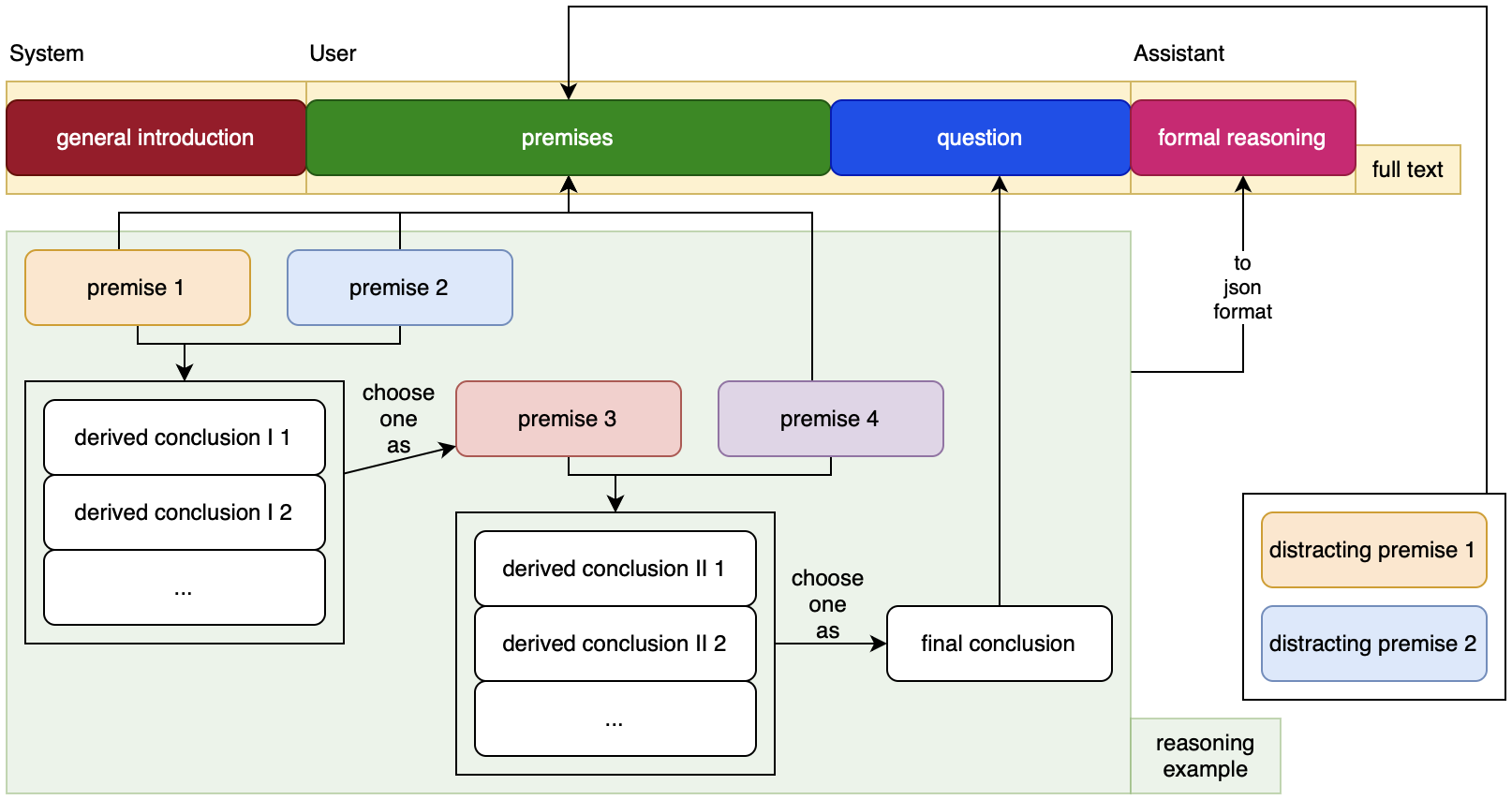}
    \caption{The overview of the prompt structure.}
    \label{fig:full prompt}
\end{figure}

We use the OpenAI User/Assistant format to generate the full prompt. The overview can be found in Fig. \ref{fig:full prompt}.

\subsection{Grading}

By requiring the LLM to output in JSON format and constraining each inference step to the minimal granularity, we can systematically evaluate whether its reasoning follows the target logical rules. Individual steps are first graded (single-step grading), followed by an assessment of their coherence (inter-step grading). When ground-truth specifying the correct conclusion(s) is available, we can further evaluate whether the target outcome is achieved (ground-truth grading).

\subsubsection{Single-Step Grading}

At the step level, we first parse the LLM’s JSON output to extract \textit{Premise 1}, \textit{Premise 2}, and \textit{Results}. If parsing fails, the minimum grade (0.1) is given (to retain gradient utility in reinforcement learning, if applicable). Otherwise, we employ the same reasoning engine used in data construction to generate all possible conclusions and identify the closest match to the LLM’s output through bipartite matching.

When calculating the matching grade, for categorical indicators ($s$, $o$, $cp$, $eb$, $r$), exact matches receive full credit (5 points), otherwise no credit (0 points). For numerical indicators ($f$, $c$), which are central to evaluating whether the LLM follows the target logic rules, grading depends on the difference from the reference value: a difference $\leq 0.05$ is considered negligible and receives full credit (25 points); a difference $\geq 0.2$ indicates category-level mistakes (since we have 5 categories defined) and receives 0 points. For intermediate cases, we apply the function $(0.1 + 0.9 \times (1 - (diff - 0.05)/0.15))^2 \times 25$. In which the squaring emphasizes penalties for larger deviations, while the 0.1 floor preserves gradient signal (for reinforcement learning). This design ensures smooth penalization within the tolerance range.

The final single-step grade is defined as the proportion of received points over the maximum possible score. This formulation also supports future extensions where a single step may yield multiple conclusions. In the following discussions, we will use the same way on matching different sentences and grade on them.

\subsubsection{Inter-Step Grading}

In multi-step reasoning, each step should build upon premises from the problem description or conclusions derived from previous steps. However, since we emphasis the NLU aspect for LLM to understand the problem and rephrase it into logic sentences, it is difficult to trace the exact origin of each premise. Therefore, we evaluate LLM outputs by analyzing their internal chain structure.

In this work, we focus on two-step reasoning tasks, where the conclusion of the first step serves as a premise in the second step, the analysis is straightforward. The same matching and grading strategy is applied on the first-step conclusion and all second-step premises.

\subsubsection{Ground-Truth Grading}

When ground-truth labels are available, we can additionally assess whether the question is correctly answered. Specifically, the same matching and grading strategy is applied on the conclusion of the second and the label answer, considering all indicators ($s$, $o$, $cp$, $f$, $c$, $eb$, $r$).

\subsubsection{Overall Evaluation}

The multiplication of the step-wise and inter-step grades define the formal conformity grade, which is standalone evaluation on whether the LLM adheres to the target logic without relying on labels. In scenarios when multiple candidate outputs are available, this grade can be used for priming structurally coherent answers (soundness). This endows the LLM with a \textit{content-agnostic evaluation} framework, so when considering answers from multiple LLMs, there is no longer a need for human inspections or for using debate among LLMs.

In testing, the formal conformity grade can further be multiplied with the ground-truth grade, yielding a comprehensive measure of whether the LLM is both logically consistent and successful in completing the reasoning task.

\subsection{Response combination}

Imagine a network of multiple LLMs where the same question may receive different answers from different models. Because their training data are highly diverse, their performance on specific problems often varies. Nevertheless, by leveraging the structured answering scheme and the previously introduced measure of logical and structural rigor, we can effectively integrate these answers.

Specifically, when different models generate reasoning steps with identical content (i.e., the same premises and conclusions), one can simply select the answer with the higher degree of formality (given whether the reasoning process provided by LLM is consistent with the target logic). When the contents diverge, single-step reasoning from different models can be combined into a new reasoning chain, thereby yielding a new answer. For example, in the two-step reasoning scenario that this paper focuses on, such composition is straightforward: the final answer can be constructed by pairing the \textit{first-step reasoning from one model} with the \textit{second-step reasoning from another}. In this case, the high-quality single steps can be picked out individually and then combined. For reasoning tasks requiring more reasoning steps, however, additional mechanisms are needed to avoid combinatorial explosion. A practical strategy is to filter out unreliable intermediate steps using formality scores, thereby reducing the scale of potential combinations.

\section{Experiment}

To evaluate our approach, we construct a network of three LLMs that solve the same task while intentionally differing in capability. The differences are induced by fine-tuning three clones of the same pretrained backbone on disjoint data subsets. Specifically, we generate two-step NAL first-order inference instances under different rules. As shown in Fig.~\ref{fig:rule table}, there are nine rule-usage patterns. We randomly partition them into three non-overlapping subsets and fine-tune each model on one subset.

All models are \textbf{fine-tuned on 100 examples}, which is relatively small dataset to show the feasibility of the work on more complicated backbones. Since larger models typically require more optimization steps to converge, we adjust the number of epochs instead of increasing data size, as follows: Qwen 2.5/0.5B with 1 epoch, Qwen 2.5/1.5B with 2 epochs, Qwen 2.5/3B with 4 epochs.

Even with more epochs, larger models often produce minor yet fatal formatting errors in their JSON outputs (e.g., unclosed brackets). We therefore apply DeepSeek R1 \cite{ahamoment} as a post-processor to repair broken JSON strings.

Although the models are tuned with different data, all three are evaluated on the same \textbf{test set of 100 examples} that uniformly covers all rules. This design is deliberate: 1) no single model is sufficient, and 2) it enables us to study the performance of priming and step-wise recomposition of intermediate reasoning steps against single-model baselines.

\subsubsection{Priming Strategies}

We consider five priming strategies for using this network of LLMs to obtain the final answer:
\begin{itemize}
    \item \textbf{Single model, no priming ($m_1$, $m_2$, $m_3$)}: use the single answer from model 1, 2, 3 as the final output.
    \item \textbf{Priming on 3 base models (model best-3, $mb_3$)}: with light priming, select the highest-scoring answer among the three models as the final output.
    \item \textbf{Step-wise recomposition (model best-9, $mb_9$)}: decompose each model’s answer into steps and recombine them across models to synthesize a new answer (3 first-step possibilities $\times$ 3 second-step possibilities).
\end{itemize}

\subsubsection{Scoring and Curves}

After priming, we evaluate each answer against the label using a grading method. Given a threshold, an answer is classified as valid or invalid. The curve reports the ratio of valid answers over the 100 testing cases. To be comprehensive, we sweep the threshold from 0.1 to 0.9 to show the performance of the LLM network under increasingly strict conditions.

\subsubsection{Experiment on Qwen2.5--0.5B}

\begin{figure}
    \centering
    \includegraphics[width=\linewidth]{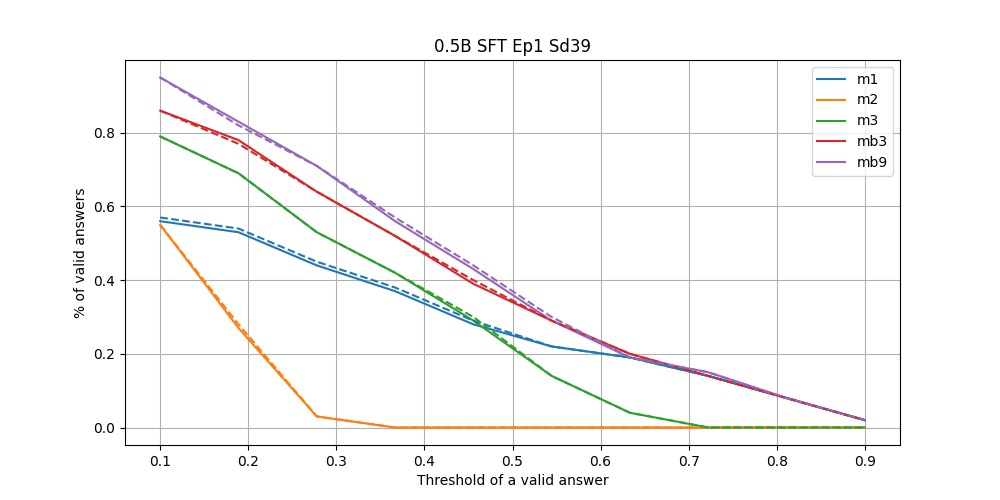}
    \caption{Ratio of valid answers under five priming strategies. Model: Qwen2.5--0.5B; fine-tuned for 1 epoch. Training set: 100 instances (seed = 39). Dashed curves denote JSON-repaired results.}
    \label{fig:q25-05b}
\end{figure}

Figure~\ref{fig:q25-05b} shows a monotonic decrease across all curves, which is expected: stricter thresholds yield lower valid-answer ratios. Among the three base models, $m_3$ performs best; however, it is still inferior to $mb_3$, indicating the effectiveness of priming in our setting. Moreover, $mb_9$ further improves over $m_3$, highlighting the additional benefit of step-wise recomposition.

The dashed curves are close to the solid ones, meaning the JSON-repair step (DeepSeek-R1 \cite{ahamoment}) has limited impact here. This suggests that small-data fine-tuning is already sufficient for the 0.5B model to produce mostly well-formatted outputs.

One potentially confusing point is why the valid-answer ratio of $m_3$ does not match that of $mb_3$, even though $mb_3$ selects the best answer among the three base outputs. This is because $mb_3$ is computed by choosing the per-example best answer \emph{before} applying the threshold, whereas $m_3$ reflects a single model’s distribution of scores; as a result, their averages need not coincide.

\subsubsection{Experiment on Qwen2.5--1.5B}

\begin{figure}
    \centering
    \includegraphics[width=\linewidth]{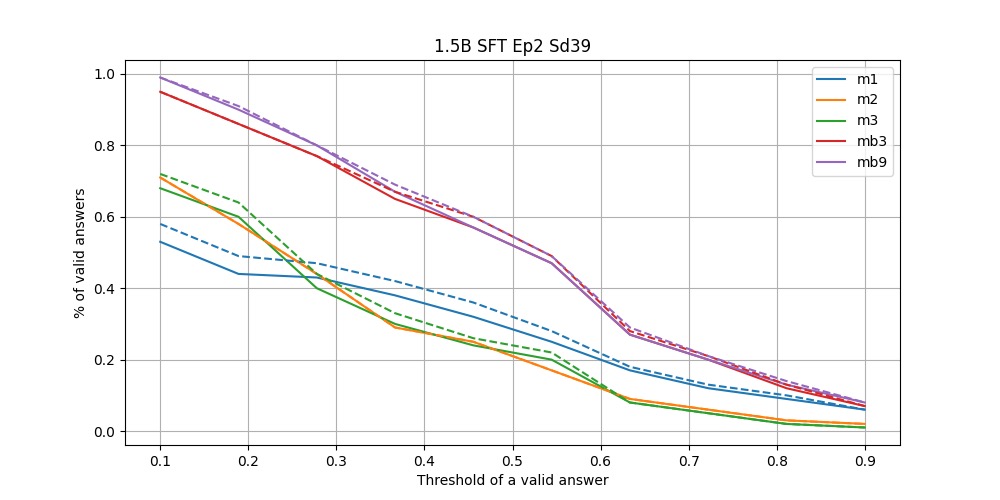}
    \caption{Ratio of valid answers under five priming strategies. Model: Qwen2.5--1.5B; fine-tuned for 2 epochs. Training set: 100 instances (seed = 39). Dashed curves denote JSON-repaired results.}
    \label{fig:q25-15b}
\end{figure}

Compared with the 0.5B results, Figure~\ref{fig:q25-15b} exhibits similar trends, with smoother curves attributable to the larger backbone. However, the dashed curves are more separated from the solid ones, underscoring the increased difficulty of enforcing correct output formatting as model size grows.

Note that within each model-size experiment, all three base models are evaluated on the same test set. Across different model-size experiments (0.5B, 1.5B, 3B), the test sets may differ.

\subsubsection{Experiment on Qwen2.5--3B}

\begin{figure}
    \centering
    \includegraphics[width=\linewidth]{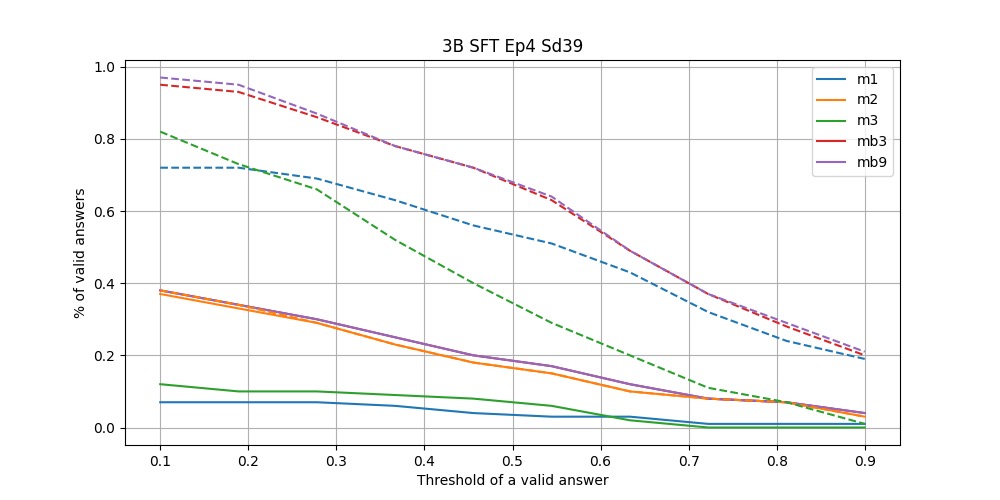}
    \caption{Ratio of valid answers under five priming strategies. Model: Qwen2.5--3B; fine-tuned for 4 epochs. Training set: 100 instances (seed = 39). Dashed curves denote JSON-repaired results.}
    \label{fig:q25-3b}
\end{figure}

In Figure~\ref{fig:q25-3b}, the dashed curve is clearly separated from the solid ones, further illustrating the challenge of maintaining strict formatting at larger scales. Meanwhile, the performance gap between $mb_3$ and $mb_9$ narrows. A plausible explanation is that the two-step task is relatively easy for the 3B model, so individual model outputs are already strong, leaving less headroom for recomposition gains.

\subsection{Comparison}

\begin{figure}
    \centering
    \includegraphics[width=\linewidth]{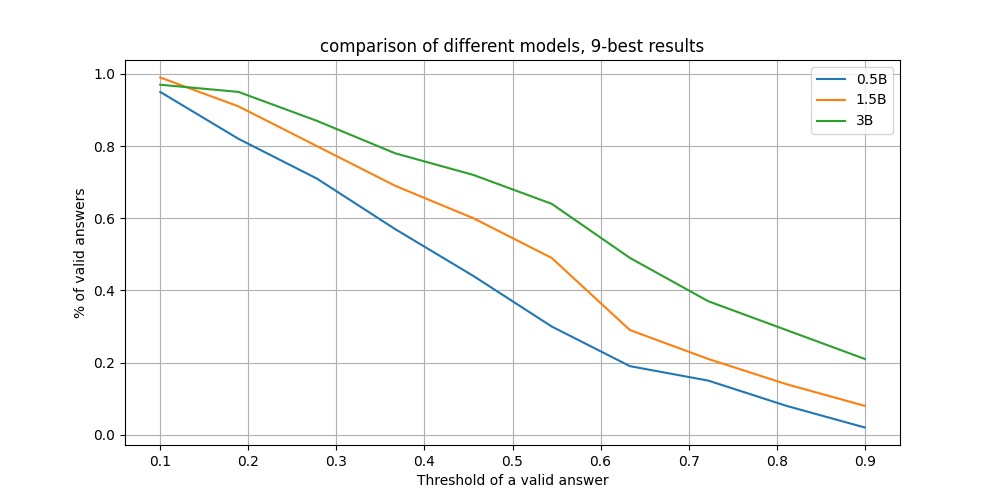}
    \caption{Comparison of the $mb_9$ valid-answer ratios across models with different parameter counts (JSON-repaired).}
    \label{fig:q25-comp}
\end{figure}

When plotting the best-performing curves (JSON-repaired $mb_9$) across parameter scales in Figure~\ref{fig:q25-comp}, we observe a clear improvement with model size. Nevertheless, even on this relatively simple two-step task, many settings still fail to exceed a valid-answer ratio of 0.2 under relatively loose thresholds, indicating substantial room for progress in faithfully simulating the target logic.

\section{Conclusion}

We proposed PCRLLM, which combines the expressive power of natural language with the rigor of formal inference rules, enabling transparent, verifiable, and collaborative reasoning with LLMs while addressing issues of trustworthiness and data scarcity.

Although this work is able to scale to more complex multi-step situations, two open issues remain for future exploration. First, since we rely on the NLU ability of LLMs, we cannot attribute which premises originate from which problem description; consequently, hallucination may still occur. While our method makes these premises transparent and allows users to accept or reject them, this workflow requires manual intervention and is therefore inconvenient.

Second, when extending to multi-step reasoning and recomposing answers from single steps, the search space grows exponentially with the depth of the NAL extensions and the number of reasoning steps, leading to a combinatorial explosion. Although our scoring scheme prunes many low-quality sub-steps and reduces fruitless search, it does not fundamentally resolve this complexity bottleneck.

Future work will focus on automating premise attribution and developing more efficient search strategies, with the aim of advancing scalable and trustworthy reasoning in LLMs.

\bibliographystyle{splncs04}
\bibliography{references}

\end{document}